\documentclass[fleqn,10pt]{wlscirep}
\usepackage[utf8]{inputenc}
\usepackage[T1]{fontenc}
\usepackage{lineno}
\usepackage{subcaption}
\usepackage[justification=centering]{caption}
\usepackage{soul}
\usepackage{color, xcolor}
\usepackage{graphicx}
\usepackage{colortbl}
\usepackage{multirow}
\usepackage{cite}
\usepackage{url}
\usepackage{hyperref}
\hypersetup{hidelinks,
    colorlinks=true,
    allcolors=blue,
    pdfstartview=Fit,
    breaklinks=true
}
\UseRawInputEncoding

\title{A Dataset with Multibeam Forward-Looking Sonar for Underwater Object Detection}

\author[1,$\dag$,*]{Kaibing Xie}
\author[1,$\dag$,*]{Jian Yang}
\author[1]{Kang Qiu}

\affil[1]{Peng Cheng Laboratory, Shenzhen, China}

\affil[*]{Kaibing Xie(xiekb@pcl.ac.cn), Jian Yang(yangj01@pcl.ac.cn)}

\begin{abstract}
\textbf{Multibeam forward-looking sonar (MFLS) plays an important role in underwater detection. There are several challenges to the research on underwater object detection with MFLS. Firstly, the research is lack of available dataset. Secondly, the sonar image, generally processed at pixel level and transformed to sector representation for the visual habits of human beings, is disadvantageous to the research in artificial intelligence (AI) areas. Towards these challenges, we present a novel dataset, the underwater acoustic target detection (UATD) dataset, consisting of over 9000 MFLS images captured using Tritech Gemini 1200ik sonar. Our dataset provides raw data of sonar images with annotation of 10 categories of target objects (cube, cylinder, tyres, etc). The data was collected from lake and shallow water. To verify the practicality of UATD, we apply the dataset to the state-of-the-art detectors and provide corresponding benchmarks for its accuracy and efficiency.}
\end{abstract}

\begin{document}

\flushbottom
\maketitle
\thispagestyle{empty}
\section*{Background \& Summary}

\hspace{1.5em}Object detection is becoming faster and more accurate with the development of AI technology. This helps underwater robots archive better performance in accident rescue, facilities maintenance, biological investigation and other underwater applications. Onshore AI algorithms are developing rapidly based on rich and high-quality datasets. In order to transfer AI achievements from land to underwater, appropriate underwater datasets are required. There have been several underwater optical datasets, such as Brackish\cite{pedersen2019detection} dataset, Segmentation of Underwater IMagery\cite{islam2020semantic} (SUIM) dataset, Detecting Underwater Objects\cite{liu2021dataset} (DUO) dataset, for the research on object detection, semantic segmentation and other AI applications. Due to the scattering and attenuation of the light in water, underwater optical imaging is a difficult task and often gets low quality images. So acoustic sensors are widely used for perceiving the underwater environment. MFLS is portable for underwater robots while providing dynamic real-time image data in high resolution. It is very applicable for scenarios requiring close and detailed inspection.\par
There have been previous works on underwater object detection and related applications with MFLS in AI areas. Haoting Zhang et al. proposed MFLS image target detection models based on You Only Look Once (YOLO) v5 network\cite{zhang2022target}. Zhimiao Fan et al. proposed a modified Mask Region Convolutional Neural Network (Mask RCNN) for MFLS image object detection and segmentation\cite{fan2021detection}. Longyu Jiang et al. proposed three simple but effective active-learning-based algorithms for MFLS image object detection\cite{jiang2020active}. Alan Preciado-Grijalva et al. investigated the potential of three self-supervised learning methods (RotNet, Denoising Autoencoders, and Jigsaw) to learn sonar image representation\cite{preciado2022self}. Gustavo Divas Karimanzira et al. proposed an underwater object detection solution with MFLS based on RCNN and deployed the solution on an NVIDIA Jetson TX2\cite{karimanzira2020object}. Neves et al. proposed a novel multi-object detection system using two novel convolutional neural network-based architectures that output object position and rotation from sonar images to support autonomous underwater vehicle (AUV) navigation\cite{neves2020rotated}. Xiang Cao et al. proposed an obstacle detection and avoidance algorithm for an AUV with MFLS, using the YOLOv3 network for obstacle detection\cite{cao2022research}. There are different defects among the datasets used in these researches, such as small sample size, few categories of objects, and virtual image data based on style transfer technology. The most important point is that most of the datasets in the related research are not public.\par
Underwater data collection often comes with high costs of economy, labor and time. Professionals are highly required in operating the MFLS devices and annotating the sonar images while most of the researchers are inexperienced in the MFLS, which results in few public related datasets\cite{choi2021physics,cerqueira2020rasterized,sung2019realistic,sung2020realistic,liu2021cyclegan}. Developing MFLS-based algorithms requires a large number of sonar images to verify and improve their approaches. For example, object detection methods in deep learning fields require data to train neural networks. Some researchers have applied sonar simulation\cite{choi2021physics,cerqueira2020rasterized} and image translation technology \cite{sung2019realistic,sung2020realistic,liu2021cyclegan} as a solution of lacking data. Since the complexity of acoustic propagation property and the instability of underwater environment, there have always been differences between the generated images and real images. Contrasting to the optical images, the lack of dataset obstructs the development of research on object detection with MFLS images in AI areas. Our proposed dataset aims to improve the above situations.\par
Considering the recognition habits of human vision, MFLS generally provides images processed with filters and pseudo-coloring which may cause the loss of effective data. Based on acoustic propagation characteristics, the MFLS provides the range and azimuth angle information. So the image in sector representation achieves better visual perception. But invalid information is imported to areas beyond the sector in the images. Figure 1 shows the MFLS raw image and the processed images. The images contain the same three target objects: ball, square cage and human body model.\par

\begin{figure}[htb]
\centering
\includegraphics[width=0.8\linewidth]{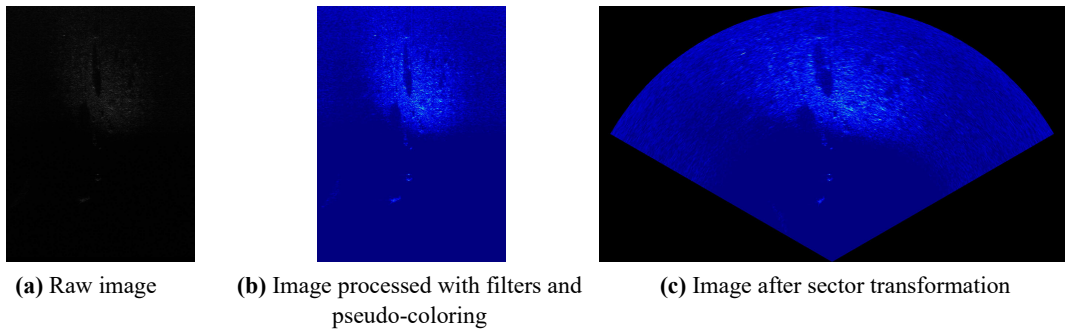}
\caption{MFLS raw and processed images}
\label{fig1}
\end{figure}

There have been several MFLS datasets providing processed image data. Erin McCann et al. provided a sonar dataset containing 8 fish species for fish classification and fishery assessment\cite{mccann2018underwater}. Deepak Singh et al. provided a sonar dataset containing typical household marine debris and distractor marine objects in 11 classes for semantic segmentation\cite{singh2021marine}. Matheus M. Dos Santos et al. published dataset ARACATI 2017, which provides optical aerial and acoustic underwater images for cross-view and cross-domain underwater localization\cite{dos2022cross}. Pontoon objects and moving boats were present in the MFLS data of the dataset. Figure 2 shows an example of comparison of the three datasets and our UATD dataset.\par

\begin{figure}[htb]
\centering
\includegraphics[width=0.8\linewidth]{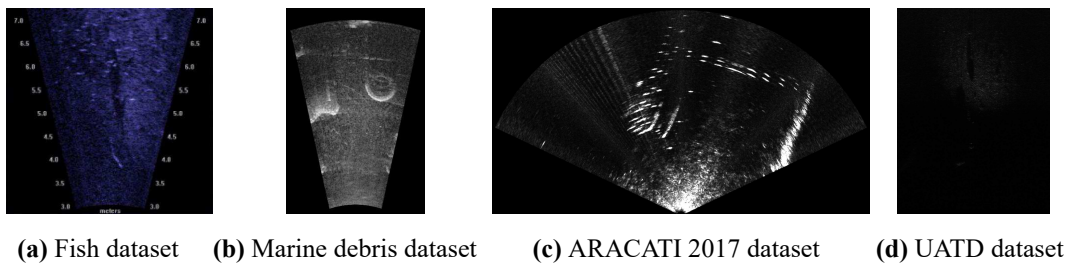}
\caption{Comparison of MFLS datasets}
\label{fig2}
\end{figure}

Our UATD dataset directly addresses the above two issues. Starting from 2020, we collected the MFLS data in Maoming and Dalian, China. The environment included lake and shallow water. The dataset provides raw data of MFLS images in high resolution with annotation of 10 categories of target objects. A corresponding benchmark of SOTA detectors performed on UATD including efficiency and accuracy indicators was provided. This dataset could promote the research on underwater object detection based on MFLS. Our work supports three consecutive China Underwater Robot Professional Contest (URPC), providing the dataset for the underwater target object detection algorithm competition. UPRC2022 refers to \href{https://challenge.datacastle.cn/v3/cmptDetail.html?id=680}{https://challenge.datacastle.cn/v3/cmptDetail.html?id=680}.\par

\section*{Methods}
\subsubsection*{Collecting MFLS Data}
\hspace{1.5em}Tritech Gemini 1200ik (website: \href{https://www.tritech.co.uk/product/gemini-1200ik}{https://www.tritech.co.uk/product/gemini-1200ik}) multibeam forward-looking sonar was used for data collection. The sonar operates at two acoustic frequencies, 720kHz for long-range target detection, and 1200kHz for enhanced high-resolution imaging at shorter ranges. Table 1 shows the acoustic specifications of the sonar. The Gemini software development kit providing the raw data of sonar images is available for Windows and Linux operating systems.\par

\begin{table}[ht]
\centering
\begin{tabular}{|l|c|c|}
     \hline
     \rowcolor{pink} Acoustic Specifications & Low Frequency Mode & High Frequency Mode \\
     \hline
     Operating frequency & 720kHz & 1200kHz \\
     \hline
     Angular resolution & 1.0$^{o}$ acoustic, 0.25$^{o}$ effective & 0.6$^{o}$ acoustic, 0.12$^{o}$ effective \\
     \hline
     Range & 0.1m-120m & 0.1m-50m \\
     \hline
     Number of beams & 512 & 1024 \\
     \hline
     Horizontal beamwidth & 120$^{o}$ & 120$^{o}$ \\
     \hline
     Vertical beamwidth & 20$^{o}$ & 12$^{o}$ \\
     \hline
     Range resolution & 4mm & 2.4mm \\
     \hline
     Update rate & \multicolumn{2}{c|}{5-65Hz(mode and range dependent)} \\
     \hline
     CHIRP support & \multicolumn{2}{c|}{Yes} \\
     \hline
     Speed of Sound & \multicolumn{2}{c|}{Integrated Velocity of Sound sensor for accuracy} \\
     \hline
\end{tabular}
\caption{\label{tab1}Acoustic specifications of Tritech Gemini 1200ik}
\end{table}

We have designed a mechanical structure for the sonar to collect data, as shown in figure 3(a). The sonar is fixed to a box structure. The box structure is mounted to the end of a metal rod. A connecting piece is installed in the middle of the metal rod to fix the collection equipment to the hull. The connecting piece allows us to adjust the rod to control the depth of sonar to the surface and the tilt angle to the water bottom during the collection. The sonar data collection structure equipped on the boat is shown in figure3(b).\par

\begin{figure}[htb]
\centering
\includegraphics[width=0.7\linewidth]{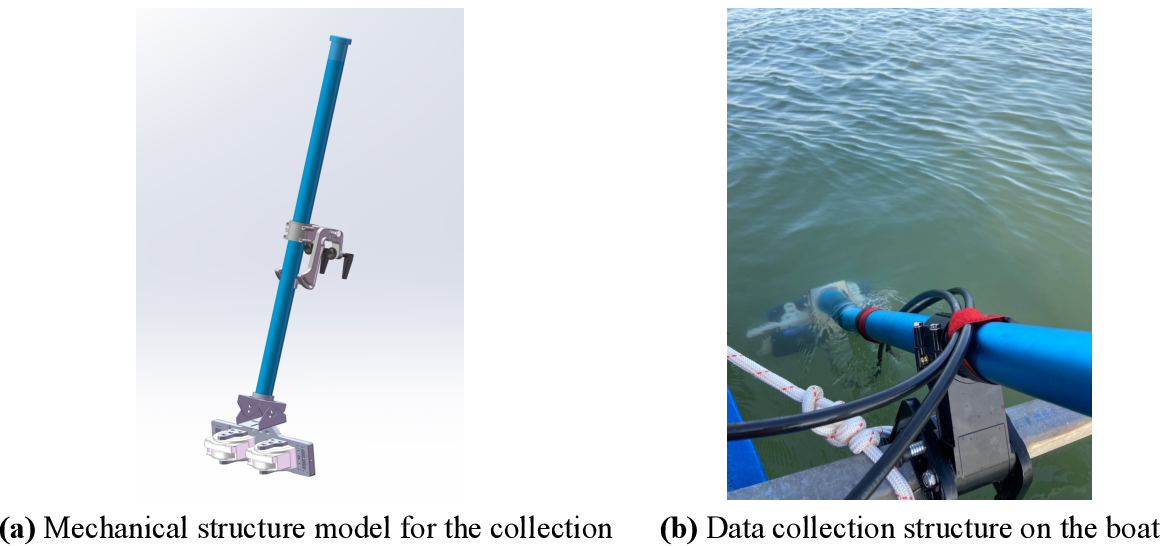}
\caption{Data collection equipment}
\label{fig3}
\end{figure}

We performed the experiments in two places: Golden Pebble Beach at Dalian(39.0904292$^{o}$N,122.0071952$^{o}$E) and Haoxin Lake at Maoming(21.7011602$^{o}$N, 110.8641811$^{o}$E). The environments of experiments performed and satellite maps with the experimental areas marked are shown in figure 4. The experimental waters have a minimum depth of 4 meters and a maximum depth of 10 meters at Dalian, and about a depth of 4 meters at Maoming.\par

\begin{figure}[htb]
\centering
\includegraphics[width=\linewidth]{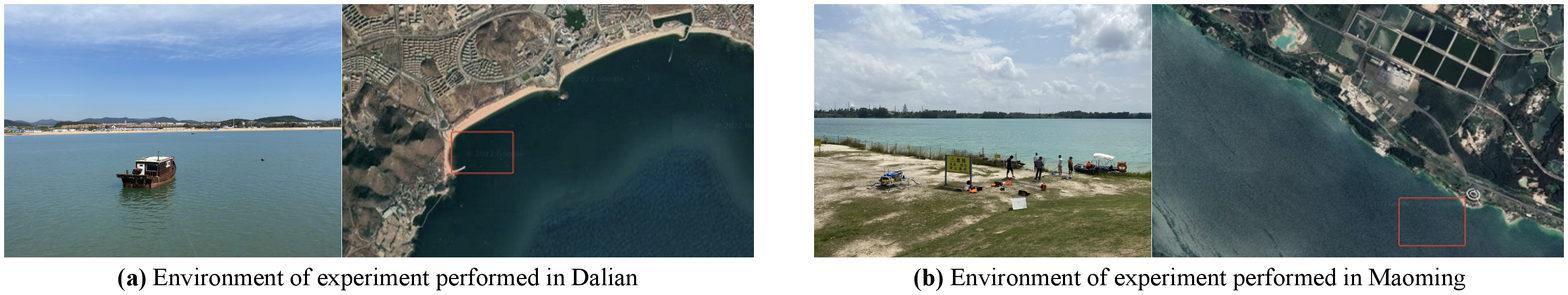}
\caption{Environment and satellite map of experiments performed}
\label{fig4}
\end{figure}

\begin{figure}[htb]
\centering
\includegraphics[width=0.8\linewidth]{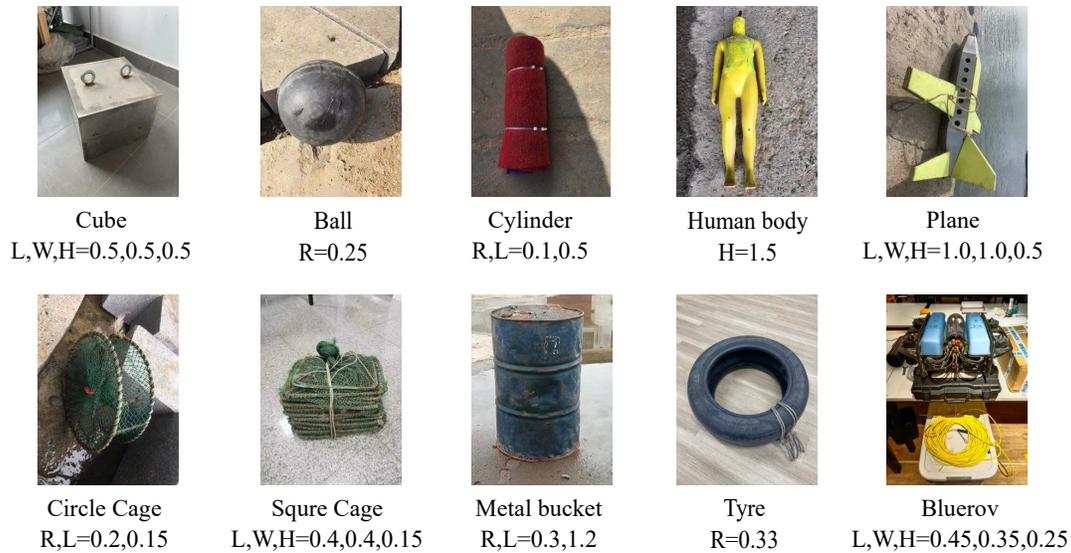}
\caption{Objects of the dataset. The scale of the objects is shown below the name of the objects in the figure. The scale is in meter and the representation: L(Length), W(Width), H(Height), R(Radius).}
\label{fig5}
\end{figure}

Ten categories of target objects were selected: cube, ball, cylinder, human body model, tyre, circle cage, square cage, metal bucket, plane model and ROV, as shown in figure 5 with their scales. The objects were tied to a floating ball with a long rope individually so that the rough location of the objects could be distinguished according to the ball floating on the water. As a result, the objects might suspend in water or lay on the bottom.\par

After deploying the objects, we drove the boat mounted with sonar and cruised around the selected sites, searching the target objects and recording data by adjusting the sonar direction.\par

\subsubsection*{Object annotations}
\hspace{1.5em}The shape of the same target may change when the sonar is imaging at different positions and angles, which makes it difficult for the annotator to judge the target category only by experience and intuition when annotating. Therefore, we developed an annotation software for sonar images named forward-looking sonar label tool (OpenSLT). Compared with other annotation tools, OpenSLT has the following two new features: 1) input as image stream. 2) real-time annotation. These features allow us to overcome the above problem as mentioned. OpenSLT can be divided into three modules: toolbar, image display area and annotation display area, as shown in figure 6(a). The tool first receives the raw data of sonar as input and plays in the form of a video stream. The playback speed can be accelerated or slowed down until the target is found. Then annotator can press the pause button and annotate in image display area using mouse. The annotation will be automatically generated and saved locally. This annotation method enables the annotator to continuously track the target object when annotating as shown in figure 6, avoiding the situations where the target object position cannot be confirmed and the target object type cannot be judged when it reappears. With the data protocol provided by Tritech, we extract the sonar working information and the sonar image information from every frame of sonar original data, including working range, frequency, azimuth, elevation, sound speed and image resolution. Then we store these information in a CSV file. OpenSLT loads the CSV file and retains these information during the annotation. In addition, OpenSLT generates file path information in the annotation for batch processing.\par

\begin{figure}[htb]
\centering
\includegraphics[width=0.9\linewidth]{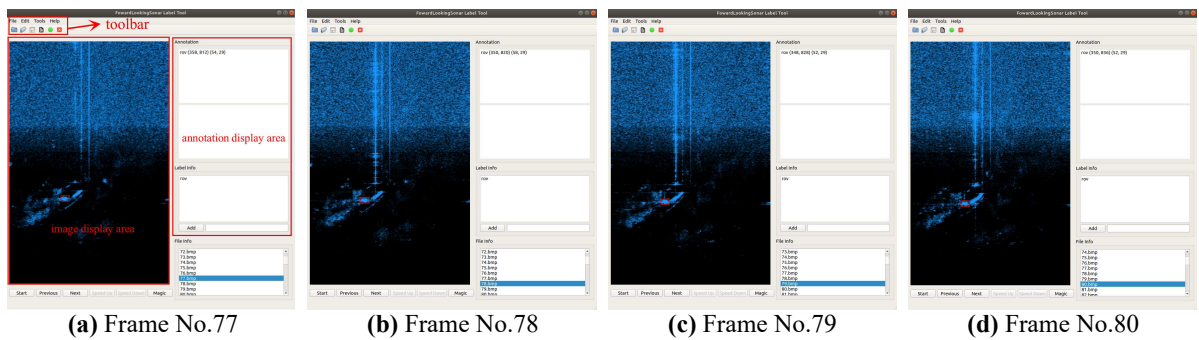}
\caption{Sequential frame annotation \\ This example shows the moving ROV during the data collection.}
\label{fig6}
\end{figure}

\section*{Data Records}
\hspace{1.5em}UATD dataset is openly available to the public in a figshare repository\cite{yangjianUATD}. The dataset contains 9200 image files in BMP format corresponding with the same number of annotation files in XML format, and is divided into three ZIP archives, namely "UATD\_Training.zip", "UATD\_Test\_1.zip", "UATD\_Test\_2.zip". "UATD-Training" contains 7600 pairs of images and annotations. The remaining two parts contain 800 pairs of images and annotations respectively. Each part consists of two folders, storing the image files and annotation files respectively. The image files are in the folder named "image", and the annotation files are in the folder named "annotation".\par
The class distribution of the objects is shown in figure 7(a). The statistic of collecting ranges and frequencies is shown in figure 7(b). A total of 2900 images have been collected in 720k Hz and 6300 have been in 1200k Hz. The sonar working range distributes from 5 meters to 25 meters during the data collection. Therefore, we count the number of images with the scope of every 1 meter by range. Besides, the images are also counted by working frequency of 720k and 1200k respectively as shown in figure 7(b).\par

\begin{figure}[ht]
\centering
\includegraphics[width=0.8\linewidth]{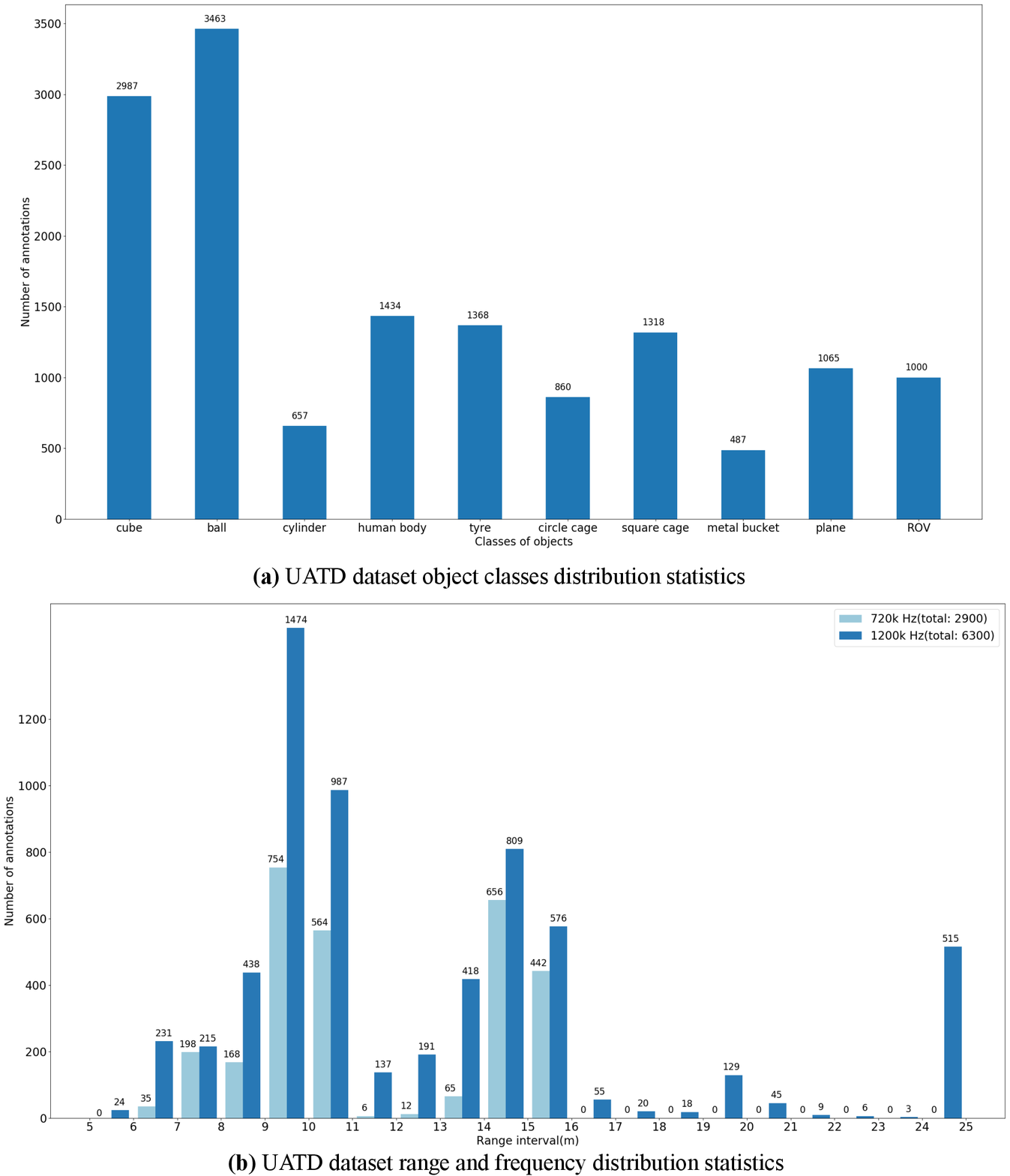}
\caption{An overview of the distribution statistics of UATD dataset}
\label{fig7}
\end{figure}

A pair of UATD data is shown in figure 8 as an example. The echo intensities data is stored in the first channel of the BMP image file. The data in the rest of the two channels of the image is the same as the first channel. The annotation file can be divided into four sections. The "sonar" section provides some basic sonar working information at the moment the corresponding image is collected. As shown in the example: the range is 14.9941m, the azimuth is 120$^{o}$, the elevation is 12$^{o}$, the sound speed is 1582.4m/s and the frequency is 1200k Hz. All of these parameters are parsed from the sonar output data stream directly. The "file" section provides some information about the relative paths of the image file and annotation file. The "filename" parameter provides the common filename prefix of a pair of image files and annotation files. The "size" section provides the image information. In this example, the image resolution is 1024x1428 and owns 3 channels. The "object" section provides the category name under "name" tab and bounding box in pixels of the object under the "bndbox" tab.\par

\begin{figure}[ht]
\centering
\includegraphics[width=0.8\linewidth]{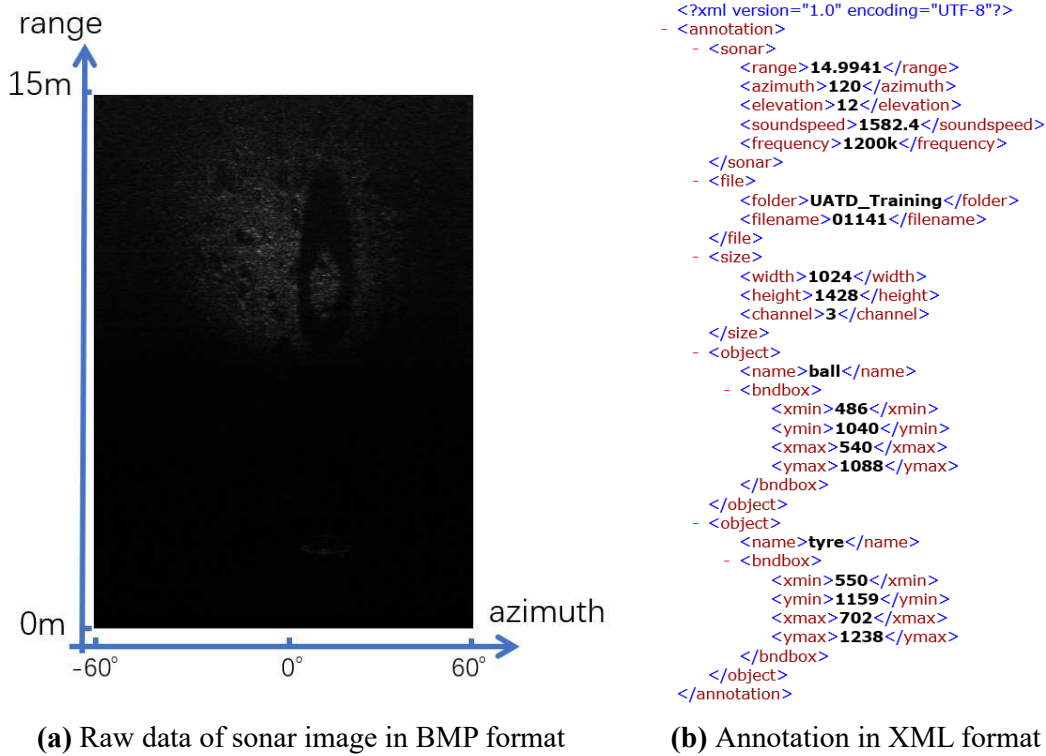}
\caption{An example of UATD dataset}
\label{fig8}
\end{figure}

\section*{Technical Validation}
\hspace{1.5em}The appearances of the same underwater target object at different imaging angles of the MFLS are generally different, leading to a great challenge to the subsequent labeling work. To address the challenge, we designed three effective methods to ensure the accuracy of sonar image annotations. Firstly, three members of our team were responsible for the annotation and completed the labeling work individually after randomly assigning the collected data. Then, with the \textit{data playback} function in OpenSLT, cross-checking was performed to reduce manual annotation errors. Secondly, we recorded the video stream displaying the processed images for vision habits from Gemini 1200ik synchronously with raw data during the data collection. At the same time, OpenSLT played back the data in the way of the stream to corresponding with the recorded video stream. So it was convenient to detect and track the object by comparing it with the video during annotating the raw data in OpenSLT, avoiding labeling errors caused by losing the target. Finally, we handed over the annotated data to a professional data management company cooperating with us. The professional staff members of the company have checked the data again to ensure correctness.
\subsubsection*{Object detection benchmarks}
\hspace{1.5em}A benchmark based on our dataset is given in table 2. Currently, MMdetection\cite{chen2019mmdetection} is one of the best open-source object detection toolboxes based on PyTorch, which provides a variety of SOTA detectors and is simple to employ. Therefore, the benchmarks are generated by MMdetection(V2.25.0). We choose Faster-RCNN\cite{ren2019towards} and YOLOv3\cite{redmon2018yolov3} with various backbones as our object detectors which are the most popular two-stage and one-stage SOTA detectors respectively.\par

\begin{table}[ht]
\centering
\setlength{\tabcolsep}{0.25mm}{
\begin{tabular}{|c|c|c|c|c|c|c|c|c|c|c|c|c|c|c|c|c|}
     \hline
     \rowcolor{pink} Model & Backbone & mAP & mAR & $AP_{ball}$ & $AP_{cube}$ & $AP_{hb}$ & $AP_{tyre}$ & $AP_{sc}$ & $AP_{plane}$ & $AP_{rov}$ & $AP_{cc}$ & $AP_{cy}$ & $AP_{mb} $ & Params & FLOPs & FPS \\
     \hline
     \multirow{3}*{Faster-RCNN} & Resnet-18 & \color{red}{0.839} & \color{red}{0.897} & 0.869 & 0.717 & 0.831 & 0.847 & 0.547 & 0.986 & 0.957 & 0.666 & 0.973 & 1.00 & 28.17M & 49.78G & 44.1 \\ ~ & Resnet-50 & 0.829 & 0.890 & 0.870 & 0.686 & 0.878 & 0.889 & 0.621 & 0.973 & 0.969 & 0.538 & 0.872 & 1.000 & 41.17M & 63.29G & 32.9 \\ ~ & Resnet-101 & 0.818 & 0.877 & 0.865 & 0.697 & 0.913 & 0.840 & 0.572 & 0.967 & 0.974 & 0.491 & 0.944 & 0.912 & 60.16M & 82.77G & 26.6 \\
     \hline
     \multirow{2}*{YOLOv3} & Darknet-53 & 0.801 & 0.880 & 0.860 & 0.669 & 0.782 & 0.874 & 0.470 & 0.988 & 0.945 & 0.519 & 0.906 & 1.000 & 61.57M & 49.67G & 49.8 \\ ~ & MobilenetV2 & 0.787 & 0.868 & 0.790 & 0.573 & 0.808 & 0.738 & 0.518 & 0.992 & 0.986 & 0.498 & 0.963 & 1.000 & \color{red}{3.68M} & \color{red}{4.22G} & \color{red}{93.4}\\
     \hline
\end{tabular}}
\caption{\label{tab2}Benchmark of Faster-RCNN and YOLOv3 with various backbones performed on UATD. (AP$_{hb}$: AP in human body, AP$_{sc}$: AP in square cage, AP$_{cc}$: AP in circle cage, AP$_{cy}$: AP in cylinder, AP$_{mb}$: AP in metal bucket)}
\end{table}

The evaluation has considered both accuracy and efficiency. We first adopted the evaluation metric mean average precision (mAP) and mean average recall rate\cite{everingham2010pascal} (mAR) to measure the accuracy of detectors on UATD. Then, the efficiency was tested on the local computer and the indicators of FPS, Params and FLOPs were given. More details as below:\par
\textbf{Accuracy metrics:}\par
\textbf{mAP} - Corresponds to the mean AP for intersect of union (IoU) equals to 0.5 on total categories (10 in UATD).\par
\textbf{mAR} - Corresponds to the mean recall rate on total categories (10 in UATD).\par
\textbf{AP$_{name}$} - AP of class (name belongs to the classes in UATD).\par
\textbf{Efficiency metrics:}\par
\textbf{Params} - The parameter size of models.\par
\textbf{FLOPs} - Floating-point operations per second with input image size of 512 x 512.\par
\textbf{FPS} - Frames per seconds.\par
The UATD was trained on the local machine with an NVIDIA GeForce GTX 1080 GPU. The image height of the input sonar image is resized to 512 while the image width is scaled by the original ratio in both training and inference. The pretrained parameters on ImageNet\cite{deng2009imagenet} were used to initialize the backbone. During the training period, the initial learning rate was set to 0.0005 and decreased by 0.1 at the 8th and 11th epoch (12 epochs in total) respectively. The warm-up strategy was adopted with a 0.0001 warm-up ratio and increased by linear step in the first 500 iterations. Otherwise, Adam method was employed to optimize the models.\par
According to the benchmark, Faster-RCNN with Resnet-18 backbone achieves the best mAP of 83.9\% and the best mAR of 89.7\%. On the other hand, YOLOv3 with MobilenetV2 backbone has a good performance in efficiency with only 3.68M Params and 4.22G FLOPs, as well as the fastest inference speed of 93.4 FPS tested on the local machine.\par

\section*{Code availability}
\hspace{1.5em}UATD dataset is published in a figshare repository\cite{yangjianUATD}. Furthermore, the annotation tool OpenSLT is published alongside the dataset, archived as "UATD\_OpenSLT.zip". OpenSLT is developed based on Qt 5.9. The tool worked well on Ubuntu 18.04/20.04 environment during our annotation work. In addition, we provide an example (a small dataset with several sonar image files and corresponding CSV files) with the tool for users to test. The file README.md along with the tool plays the role of guidance for the users. 

\section*{Acknowledgements}
\hspace{1.5em}This work was supported by the National Natural Science Foundation of China (Grant, No.62027826). We thank the Dalian Key Laboratory of Underwater Robot of Dalian University of Technology for their support during the data collection. We thank the support provided by OpenI Community(\href{https://openi.pcl.ac.cn}{https://openi.pcl.ac.cn}) during data processing.

\section*{Author contributions}
\hspace{1.5em}Kang Qiu, Jian Yang and Kaibing Xie generated the dataset. Jian Yang contributed to the annotation software development and technical validation. Kaibing Xie draft the paper and provided feedback on the manuscript.

\section*{Competing interests}
\hspace{1.5em}The authors declare no competing interests.

\bibliography{reference}

\end{document}